\pdfoutput=1
\documentclass{article}

\usepackage[preprint]{corl_2026} % Uncomment for pre-prints (e.g., arxiv); This is like ``final'', but will remove the CORL footnote.
\usepackage{amsmath}
\usepackage{amsfonts}
\usepackage{amssymb}
\usepackage[table]{xcolor}
\selectcolormodel{rgb}% reduce ``Incompatible color definition'' warnings with corl's legacy \RequirePackage{color}
\definecolor{lightblue}{RGB}{230,240,255}
\definecolor{lightgreen}{RGB}{230,255,235}
\definecolor{lightyellow}{RGB}{255,250,230}
\definecolor{highlight}{RGB}{200,230,200}
\definecolor{revisionblue}{RGB}{31,119,180}
\definecolor{revisionred}{RGB}{214,39,40}
\definecolor{revisionorange}{RGB}{255,127,14}

\usepackage{algorithm}
\usepackage{algorithmic}
\usepackage{booktabs}
\usepackage{array}
\usepackage{makecell}
\usepackage{tabularx}
\usepackage{multirow}
\usepackage{colortbl}
\usepackage{wrapfig}
\newcolumntype{Y}{>{\raggedright\arraybackslash}X}
\usepackage{tikz}
\usepackage{pifont}
\newcommand{\cmark}{\ding{51}}
\newcommand{\xmark}{\ding{55}}
\usetikzlibrary{arrows.meta,positioning,shapes.geometric,calc}

\title{UniReflex: Plug-and-Play Force Control for Pretrained Generative Policies via Fast-Slow Reflex}

\author{
  \normalfont
  \begin{minipage}{0.98\textwidth}
    \centering
    \small
    \textbf{Yan Huang}$^{1,\dagger}$ \quad
    \textbf{Shoujie Li}$^{2,\dagger}$ \quad
    \textbf{Ziwu Song}$^{1,3}$ \quad
    \textbf{Wenbo Ding}$^{1,4,*}$\\[3pt]
    $^{1}$Tsinghua University Shenzhen International Graduate School\\
    $^{2}$Nanyang Technological University \quad
    $^{3}$X Square Robot \quad
    $^{4}$Xspark AI\\
    \footnotesize
    \url{https://unireflex.github.io/}\\
    $^{\dagger}$Equal contribution \quad
    $^{*}$Corresponding author
  \end{minipage}
}

\begin{document}
\maketitle

% ====== Main paper content ======
%===============================================================================

\begin{abstract}
Generative imitation learning policies excel at trajectory planning but lack closed-loop force regulation, while directly incorporating force modalities often requires redesigning or retraining the network.
We present UniReflex, a universal plug-and-play framework that equips frozen generative policies with variable impedance control (VIC) for contact regulation, guided by force-direction intent collected during demonstration, without further slow-backbone fine-tuning.
By non-invasively intercepting deep latent representations from the action head, UniReflex drives a fast reflex network that decouples active force exertion from external interaction response.
This scheme predicts normalized anisotropic stiffness directions for directional compliance allocation.
Furthermore, UniReflex integrates an adaptive gating mechanism that enables seamless transitions between position-dominant planning and force-dominant execution.
Real-world bimanual experiments demonstrate that UniReflex significantly improves contact stability and success rates while preserving original position accuracy.
Our approach achieves $25$--$66\times$ lower per-step backward latency relative to joint training strategies on the evaluated backbones.
\end{abstract}


%===============================================================================

\section{Introduction}
\label{sec:intro}

Imitation learning has become the dominant recipe for scalable visuomotor manipulation: Diffusion Policy (DP)~\cite{chi2025diffusion}, vision-language-action (VLA) models~\cite{black2026pi0visionlanguageactionflowmodel,kim2024openvlaopensourcevisionlanguageactionmodel}, and world action models (WAM) that learn predictive dynamics or video-grounded latents before decoding actions~\cite{ye2026worldactionmodelszeroshot} now reach strong performance on diverse tabletop and pick-and-place regimes~\cite{mandlekar2021matterslearningofflinehuman}.
Yet contact-rich, force-sensitive skills, such as wiping, insertion, peeling, and sustained pressing, often remain unreliable in deployed systems~\cite{doi:10.1177/02783649261417694}.
The dominant recipe predicts kinematic motion targets at the policy level and tracks them with approximately fixed low-level gains; under this pattern, small variations in geometry, compliance, or disturbance can destabilize contact in ways that are difficult to remedy through data scaling alone.
This fundamental limitation highlights a gap between purely geometric motion planning and the physical realities of robotic deployment.

In parallel, a growing line of work targets contact by making forces explicit in learning and control, e.g., force-conditioned reactive policies~\cite{he2024foar}, force-aware VLAs~\cite{NEURIPS2025_8633b46e}, hybrid force-position formulations~\cite{fang2026forcepolicylearninghybrid,li2026mastermicroresidualcorrection}, and tactile or vision-driven fast-slow policies~\cite{xue2025reactive}.
These advances show that wrench and tactile cues can materially improve manipulation, but they also diversify how force information is used.
By directly incorporating physical interaction data, these methods attempt to bridge the gap between visual intent and tactile execution.

Concretely, recent methods cluster into three families.
First, many approaches treat force or torque primarily as additional observations: force/torque or tactile streams are fused into the policy input while the network still predicts trajectory-like actions~\cite{he2024foar,NEURIPS2025_8633b46e,li2026biomimetic,10175024}.
Second, another family couples visuomotor policies with hybrid force-position or variable impedance control (VIC), explicitly regulating some degrees of freedom in force and others in pose~\cite{fang2026forcepolicylearninghybrid,hou2025adaptive}.
Third, fast-slow hierarchies decouple slow visual reasoning from high-rate contact feedback~\cite{xue2025reactive}, e.g., via tactile reactive heads or cascaded decoders~\cite{bu2025synergisticgeneralizedefficientdualsystem}.
Across these families, however, a recurring practical cost is structural: realizing their full pipelines typically requires end-to-end training, architectural refits, or joint optimization. 
This poses a significant dilemma in the era of foundation models. Today's most capable generative policies are pre-trained on massive, predominantly vision-only datasets that lack explicit force information. 
Discarding these powerful pre-trained weights---or subjecting them to heavy fine-tuning to accommodate new force modalities---is highly inefficient and risks destroying their well-generalized distributions. 
Because force feedback is a specialized and relatively sparse observation, it should ideally be integrated as an independent module. 
A principled solution must therefore preserve the original pre-trained distribution of the foundation model while independently endowing the system with robust physical reflexes.

This raises a direct question: can we add closed-loop contact regulation as a \textbf{lightweight, plug-and-play module} without further fine-tuning the already task-adapted slow policy, while still benefiting from its high-level intent?
We present \textbf{UniReflex}, a general plug-and-play framework that attaches fast contact regulation to frozen generative policies through a decoupled design for denoising/generative backbones: no model-specific fast-slow interface is required, and only the lightweight reflex is trained while the backbone stays frozen.
We further verify that directly reusing the latent action representation from the action generation head provides a recoverable representation for downstream force control, and that the same reflex interface transfers across heterogeneous generative policies.
By doing so, we aim to offer a scalable alternative to monolithic end-to-end training pipelines.
Table~\ref{tab:method_comparison} provides a compact, deployment-oriented comparison of representative methods.

\noindent Our contributions are threefold:
\begin{enumerate}
    \item \textbf{A Universal Plug-and-Play Framework:} UniReflex equips supervised fine-tuned generative policies with closed-loop force control by non-invasively intercepting deep action-head latent representations, without additional training or fine-tuning of the slow backbone.
    \item \textbf{Force-Stiffness Decoupling:} We jointly supervise normalized anisotropic stiffness directions and reference forces, enabling VIC-based closed-loop regulation that yields more physically consistent wrench behavior.
    \item \textbf{Adaptive Gating for Hybrid Control:} A lightweight gating mechanism switches between position-dominant and force-dominant execution so the robot preserves the frozen planner's alignment quality while gaining robust regulation during contact.
\end{enumerate}

% === Comprehensive comparison table (table* must not be inside \colorbox/\parbox) ===
\begin{table*}[t]
  \centering
  \caption{\textbf{Taxonomy of force control methods.} 
  UniReflex uniquely enables active, variable-stiffness control while maintaining plug-and-play compatibility with frozen generative policies via fast-only training.}
  \label{tab:method_comparison}
  \begingroup
  \small
  \setlength{\tabcolsep}{3.5pt}
  \renewcommand{\arraystretch}{1.15}
  % Vertically center text columns (m / X-as-m) so header aligns with \shortstack in c columns (p/X default to top alignment).
  \renewcommand{\tabularxcolumn}[1]{>{\raggedright\arraybackslash}m{#1}}
  % Last six columns use fixed-width m{...} (vertically centered) so header \shortstack aligns with left m/X columns; plain c + p/m mixes top vs mid alignment.
  \begin{tabularx}{\textwidth}{@{}>{\raggedright\arraybackslash}m{3.35cm} Y *{6}{>{\centering\arraybackslash}m{1.14cm}}@{}}
  \toprule
  \textbf{Category} & \textbf{Representative methods} & \makecell{\textbf{Force}\\\textbf{ctrl.}} & \makecell{\textbf{Closed-}\\\textbf{loop}} & \makecell{\textbf{Variable}\\\textbf{stiff.}} & \makecell{\textbf{Ref.}\\\textbf{force}} & \makecell{\textbf{Plug-}\\\textbf{play}} & \makecell{\textbf{Fast-only}\\\textbf{training}} \\
  \midrule
  \multicolumn{2}{@{}l}{\textbf{Variable impedance control}} & & & & & & \\
  \addlinespace[1pt]
  Guided Control & ACP~\cite{hou2025adaptive} & Active & \cmark & \cmark & \cmark & \xmark & \xmark \\
  Hybrid P/F Control & ForceVLA2~\cite{li2026forcevla2unleashinghybridforceposition} & Active & \cmark & \cmark & \cmark & \xmark & \xmark \\
  % Online adaptation & Kronander et~al.~\cite{kronander2015onlinevic} & Passive & \cmark & \cmark & \xmark & \xmark & \xmark \\
  \midrule
  \multicolumn{2}{@{}l}{\textbf{Force-conditioned imitation learning}} & & & & & & \\
  \addlinespace[1pt]
  Force-aware DP & FoAR~\cite{he2024foar} & Passive & \xmark & \xmark & \xmark & \xmark & \xmark \\
  Force-aware VLA & TA-VLA~\cite{zhang2025elucidating}, ForceVLA~\cite{NEURIPS2025_8633b46e} & Passive & \xmark & \xmark & \xmark & \xmark & \xmark \\
  \midrule
  \multicolumn{2}{@{}l}{\textbf{Fast-slow systems}} & & & & & & \\
  \addlinespace[1pt]
  Tactile-reactive DP & RDP~\cite{xue2025reactive} & Passive & \cmark & \xmark & \xmark & \xmark & \xmark \\
  VLA-diffusion hybrid & RoboDual~\cite{bu2025synergisticgeneralizedefficientdualsystem} & Passive & \xmark & \xmark & \xmark & \cmark & \cmark \\
  \midrule
  \rowcolor[rgb]{0.99,0.96,0.78}
  \textbf{UniReflex (Ours)} & \textbf{-} & \textbf{Active} & \textbf{\cmark} & \textbf{\cmark} & \textbf{\cmark} & \textbf{\cmark} & \textbf{\cmark} \\
  \bottomrule
  \end{tabularx}
  \endgroup
  \end{table*}

\section{Related Work}
\label{sec:related}

\noindent
We organize concurrent work along the same axes as the Introduction: high-capacity imitation policies, force-aware learning and control, and fast-slow decompositions.
Table~\ref{tab:method_comparison} provides a compact architectural comparison based on strict operational criteria: \textbf{Active} force control explicitly commands force/stiffness targets, whereas \textbf{Passive} relies purely on kinematics; \textbf{Closed-loop} denotes real-time action adjustment specifically via force, torque, current, or tactile feedback; \textbf{Variable stiff.} and \textbf{Ref. force} indicate explicit prediction of stiffness parameters and desired force trajectories; \textbf{Plug-play} and \textbf{Fast-only} highlight the ability to avoid fine-tuning the frozen slow backbone. Here we summarize each thread and how UniReflex differs in problem setting.

\subsection{Generative Imitation Learning for Manipulation}
Modern generative imitation policies, including diffusion trajectory models~\cite{chi2025diffusion,Ze2024DP3}, CVAE-style sequence models~\cite{zhao2023learningfinegrainedbimanualmanipulation}, VLAs~\cite{openxe2023rtx,black2026pi0visionlanguageactionflowmodel,kim2024openvlaopensourcevisionlanguageactionmodel,li2024cogactfoundationalvisionlanguageactionmodel}, and WAM~\cite{ye2026worldactionmodelszeroshot}, achieve strong visuomotor performance on diverse manipulation benchmarks~\cite{mandlekar2021matterslearningofflinehuman}.
Efficient fine-tuning recipes remain the default path for adapting large VLAs to new robots~\cite{kim2025finetuningvisionlanguageactionmodelsoptimizing}, and dual-encoder strategies that preserve frozen pretrained vision--language representations while adapting robot-specific pathways have been proposed to mitigate catastrophic forgetting during VLA adaptation~\cite{grover2025enhancinggeneralizationvisionlanguageactionmodels}.
Their standard interfaces nevertheless remain dominated by kinematic action prediction with fixed-gain tracking, so contact wrenches are rarely regulated as explicit closed-loop objectives at deployment.

\subsection{Force-Aware Robot Control}
Classical and learning-based impedance and variable-stiffness control~\cite{hogan1985impedance,doi:10.1177/0278364911402527,10.3389/frobt.2020.590681,zhang2024learningvariableimpedanceskills} shape compliant contact by modulating stiffness and damping over time.
Generative models have also been used to learn variable impedance and contact-consistent equilibria online~\cite{10801976,geiger2026diffusionbasedimpedancelearningcontactrich}.
Recent learning systems further couple visuomotor policies with hybrid force-position execution~\cite{hou2025adaptive,NEURIPS2025_8633b46e,fang2026forcepolicylearninghybrid,li2026forcevla2unleashinghybridforceposition,ge2025filicdualloopforceguidedimitation,li2026mastermicroresidualcorrection}.
In parallel, force-conditioned imitation learning incorporates force/torque or tactile streams as observations~\cite{he2024foar,11128061,zhang2025elucidating,NEURIPS2025_8633b46e,lee2025manipforceforceguidedpolicylearning,10752344,li2026biomimetic,10175024}, improving contact-rich behavior through representation learning even when the deployed controller is not a classical hybrid force-position controller.
Across these families, realizing their full pipelines typically requires end-to-end training or coordinated fine-tuning of perception and action, rather than composing a post-hoc module atop an already trained generative policy.

\subsection{Hierarchical Fast-Slow Policies}
Fast-slow and dual-rate hierarchies address the latency gap between slow VLA inference and millisecond-scale contact by cascading decoders, tactile heads, or asynchronous buffers~\cite{xue2025reactive,bu2025synergisticgeneralizedefficientdualsystem,chen2025fastinslowdualsystemfoundationmodel,han2024dualprocessvlaefficient,zou2025asynchronousfastslowvisionlanguageactionpolicies,song2025humeintroducingsystem2thinking,cui2025openhelixshortsurveyempirical,li2026favlaforceadaptivefastslowvla,11128816,zhao2025touchbeginsvisionends}.
These designs improve temporal alignment for contact, but prior art typically \emph{co-trains} slow and fast modules, which is expensive for large backbones and misaligned with upgrading frozen, field-deployed policies.
UniReflex instead keeps the slow generative policy frozen, trains only a compact reflex on proprioceptive feedback and hooked action-head features, and targets active variable-stiffness wrench regulation.

%===============================================================================
\section{Preliminaries}
\label{sec:prelim}

\begin{figure}[t]
  \centering
  \includegraphics[width=\linewidth]{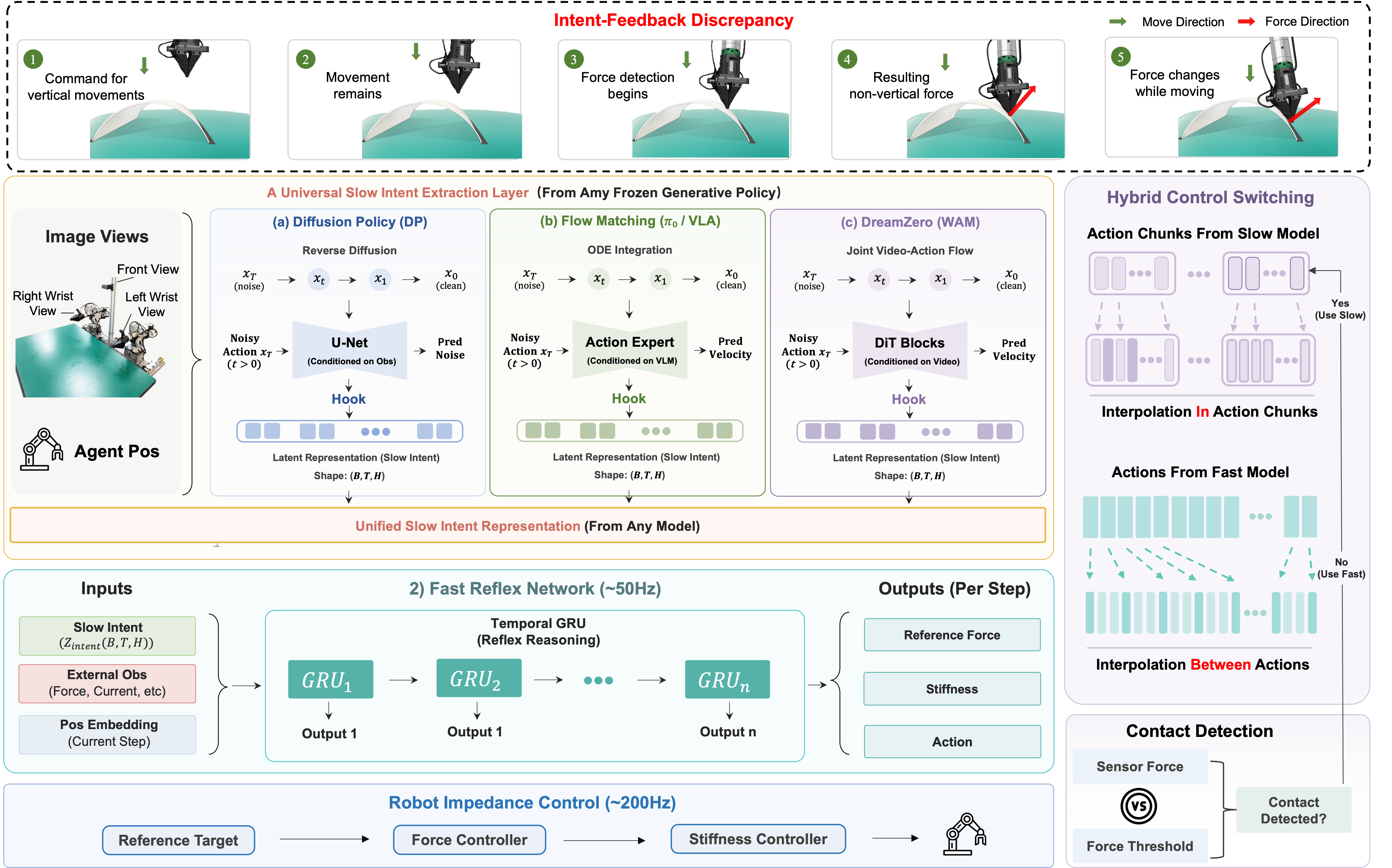}
  \caption{\textbf{System architecture of UniReflex}. UniReflex hierarchically couples slow vision-driven planning, a fast proprioceptive reflex, and a real-time execution layer that tracks predicted reference forces together with reference-impedance targets.}
  \label{fig:system_overview}
\end{figure}

\subsection{Generative Action Heads}
\label{sec:prelim_denoising}

Modern generative policies differ in objectives and architectures: DP~\cite{chi2025diffusion} predicts noise ($\epsilon$), while VLAs like $\pi_0$~\cite{black2026pi0visionlanguageactionflowmodel} and WAMs like DreamZero~\cite{ye2026worldactionmodelszeroshot} predict flow velocities ($v_t$). Despite this, their action heads share a core pattern: decoding actions via iterative denoising or flow integration. We unify these paradigms at their final generation step ($k \to 0$). Here, the internal state closely approximates the clean action sequence $x_0$. To predict the final tiny residual or velocity, the network must encode a comprehensive trajectory understanding within its hidden layers. Before the final projection collapses this into low-dimensional kinematics, models maintain a temporally structured hidden representation. We reshape this pre-projection state as $z\in\mathbb{R}^{B\times T_{\mathrm{slow}}\times H}$ (batch, action horizon, hidden width). UniReflex intercepts this as a backend-agnostic ``slow intent'' interface, preserving task semantics without architectural surgery or slow-policy fine-tuning.

% \paragraph{A fundamental physical limitation.}
% At the same time, objectives such as \eqref{eq:prelim_dp_loss} and \eqref{eq:prelim_cfm_loss} penalize mismatch primarily on a \emph{spatial} action manifold and do not introduce an explicit penalty on contact wrenches (measured interaction forces/torques).
% Consequently, when deployed in contact-rich settings, pretrained policies behave largely like \emph{open-loop} Cartesian planners: they can ``fight'' the environment through stiff tracking rather than regulating sustained contact forces.
% Closing this gap requires low-level \emph{closed-loop} physical correction compatible with the frozen generative stack.

\subsection{Variable Impedance Control}
\label{sec:prelim_contact}

At execution time, contact-rich manipulation is often closed with variable impedance control (VIC): the end-effector is regulated as a \emph{virtual} mass--spring--damper instead of as a stiff pose tracker.
In Cartesian coordinates, a standard error-based model is
\begin{equation}
  \Lambda \ddot{\tilde{\mathbf{x}}}
  \;+\;
  D(t)\,\dot{\tilde{\mathbf{x}}}
  \;+\;
  K(t)\,\tilde{\mathbf{x}}
  \;=\;
  \mathbf{F}_{\mathrm{ext}},
  \label{eq:prelim_vic}
\end{equation}
where $\tilde{\mathbf{x}}$ is end-effector pose error, $\mathbf{F}_{\mathrm{ext}}$ is the contact wrench measured at the wrist, $K(t)$ sets how stiff the arm is along each axis, and $\Lambda$ fixes the apparent end-effector inertia.
In practice, the time-varying gains $K(t)$ and $D(t)$ are commonly scheduled online from contact measurements together with task-phase logic~\cite{geiger2026diffusionbasedimpedancelearningcontactrich}.

%===============================================================================
\section{Method}
\label{sec:method}

\subsection{Active Effort Recovery and Variable-Stiffness Labeling}
\label{subsec:effort_stiffness_labels}

We build supervision signals from teleoperation logs.
Consider an $n$-degree-of-freedom bilateral teleoperation stack whose master and slave arms obey
\begin{align}
  M_m(q_m)\ddot{q}_m + C_m(q_m,\dot{q}_m)\dot{q}_m + G_m(q_m) &= \tau_m + \tau_h,\\
  M_s(q_s)\ddot{q}_s + C_s(q_s,\dot{q}_s)\dot{q}_s + G_s(q_s) &= \tau_s + \tau_e,
  \label{eq:teleop_dynamics}
\end{align}
where $(q_m,q_s)$ are joint coordinates, $(\tau_h,\tau_e)$ are interaction torques, and quasi-static motion with stiff slave tracking $\tau_s = K_p(q_m-q_s)+K_d(\dot{q}_m-\dot{q}_s)+G_s(q_s)$ implies $\tau_e \approx K_p(q_m-q_s)$; mapping this relation to Cartesian end-effector coordinates yields a loggable proxy for interaction-force intent,
\begin{equation}
  F_{\mathrm{intent}} \approx K_{\mathrm{virtual}}\,\Delta x,
  \qquad \Delta x = x_m-x_s,
  \label{eq:f_intent_virtual}
\end{equation}
where $K_{\mathrm{virtual}}$ is the effective Cartesian stiffness map implied by the slave position loop and the kinematic Jacobian.

In physical interaction, stiffness should rise with commanded effort rather than relying on stiffness reduction alone for compliance.
We apply this qualitative rule to demonstrations: stronger mismatch cues along the teleoperation error direction justify a stiffer axis in the supervised impedance targets, improving contact stability with wrist force/torque sensing only (no extra hybrid force-position hardware).

Let $n=\Delta x/\|\Delta x\| \in \mathbb{R}^3$ be the principal effort axis from teleoperation mismatch.
To ensure stability on standard industrial controllers, we project this intent into a decoupled diagonal stiffness template,
\begin{equation}
  K_{\mathrm{label}} = \mathrm{diag}(|n_x|, |n_y|, |n_z|) + \epsilon I,
  \qquad \epsilon>0,
  \label{eq:k_label_rank1}
\end{equation}
which encodes anisotropic stiffness magnitudes while simplifying the learning objective and guaranteeing positive-definiteness.

\subsection{Non-Invasive Internal Representation Interception}
\label{sec:latent_interception}

Building on the shared action-head structure in Section~\ref{sec:prelim_denoising}, we read multimodal intent from \emph{late} internal activations without rewiring the vision stack. Crucially, we observe that whether the generative mechanism is diffusion (DP) or flow matching (VLA and WAM), the policy invariably forms a temporally structured hidden representation prior to final action projection. We formalize this intercepted feature as a unified latent representation, which encapsulates the slow policy's intent:
\begin{equation}
  z = \mathrm{Hook}(\pi_{\mathrm{gen}}, o_t) \in \mathbb{R}^{B \times T_{\mathrm{slow}} \times H},
  \label{eq:z_intercept}
\end{equation}
where $B$ is the batch size, $T_{\mathrm{slow}}$ is the action horizon, and $H$ is the hidden dimension.

Concretely, the noise (or flow) predictor admits a standard decomposition,
\begin{equation}
  \epsilon_\theta(x_k, o_t, k) = W_{\mathrm{out}}\,\Phi_{\mathrm{dec}}(x_k, o_t, k),
  \label{eq:dec_decomp}
\end{equation}
where $\Phi_{\mathrm{dec}}$ aggregates deep multimodal reasoning and $W_{\mathrm{out}}$ maps to action coordinates. Near the end of reverse synthesis ($k\!\to\!0$), the sampled trajectory geometry is largely committed; we register a \emph{forward} hook on $\Phi_{\mathrm{dec}}$ to extract $z$. The hook is read-only with respect to frozen backbone weights, so the slow policy needs neither architectural surgery nor joint fine-tuning. This unified $\mathbb{R}^{B \times T_{\mathrm{slow}} \times H}$ interface allows UniReflex to act as a backend-agnostic module across heterogeneous generative policies.

\begin{figure}[t]
  \centering
  \includegraphics[width=1.00\linewidth]{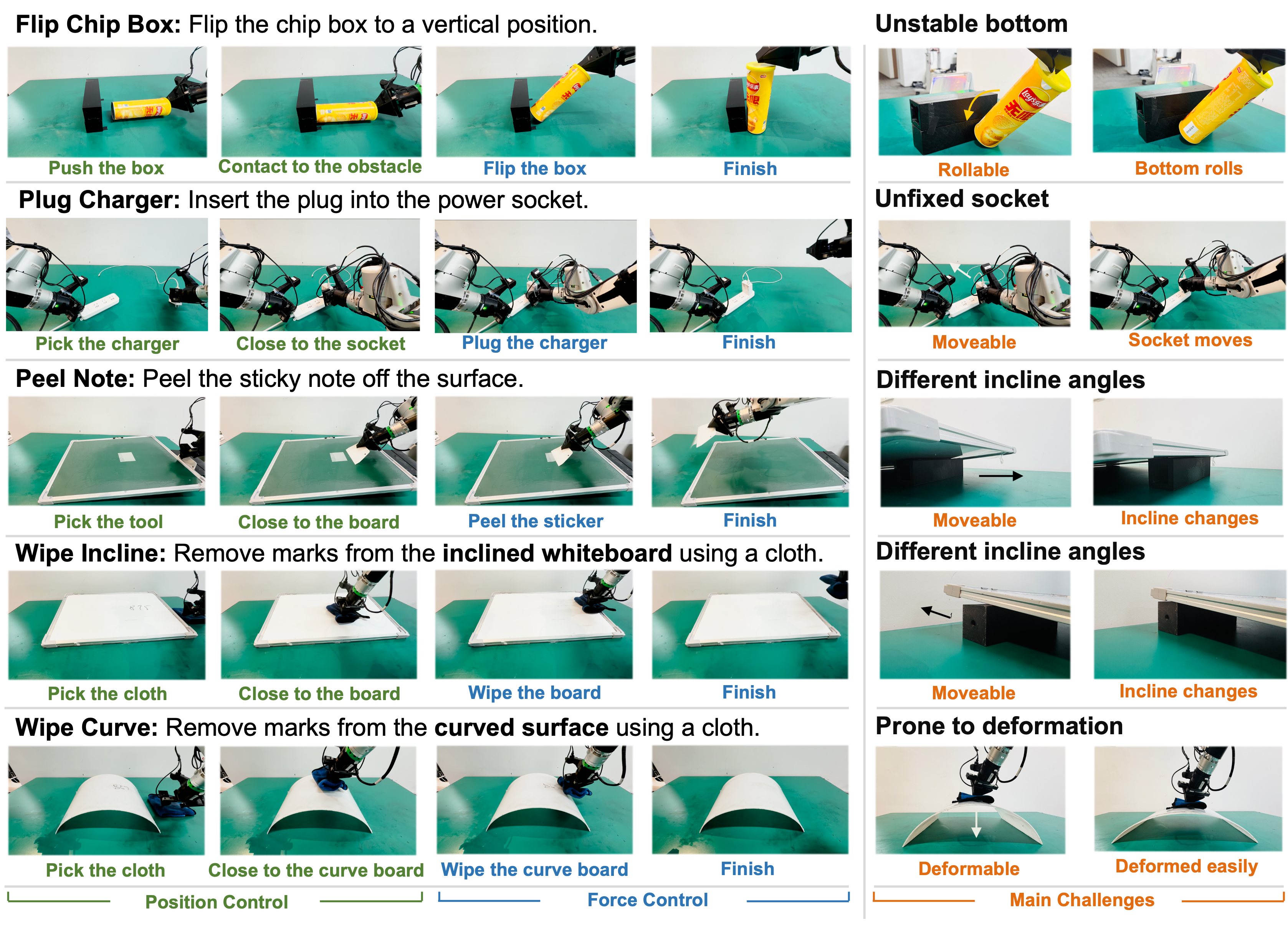}
  \caption{\textbf{Five contact-rich tasks:} short per-task introductions and the main physical challenges, together with a two-stage evaluation protocol.}
  \label{fig:tasks_overview}
\end{figure}

\subsection{Intent-Conditioned State Observation and Reflex Decoding}
\label{subsec:recurrent_decoding}

Many generative policies use action chunking, predicting a $T_{\mathrm{slow}}$-step horizon $A_{\mathrm{chunk}}=[x_t,\ldots,x_{t+T_{\mathrm{slow}}-1}]$ in one pass; under contact, open-loop rollout of long chunks reacts poorly to impacts and friction.
The reflex $\phi_{\mathrm{reflex}}$ therefore emits a single-step update each control cycle; we employ a causal temporal model so short-term contact dynamics are retained in memory.
At time~$t$, the fast network receives real-time extended observations $e_t$ and the unified latent representation $z \in \mathbb{R}^{B \times T_{\mathrm{slow}} \times H}$ intercepted from $\pi_{\mathrm{gen}}$, which serves as the slow intent.

To bridge the temporal gap between the slow chunk and the high-frequency reflex, we first temporally align the latent representation $z$ to the fast action horizon, yielding $z'_t$.
At each fast step $t$, the input $u_t$ is constructed by concatenating three encoded features: the step-wise slow intent condition $s_t = f_{\mathrm{slow}}(z'_t)$, the real-time feedback $r_t = f_{\mathrm{ext}}(e_t)$, and a learnable step positional embedding $p_t = \mathrm{Emb}(t)$ that indicates the current phase within the chunk,
\begin{equation}
  u_t = \bigl[ f_{\mathrm{slow}}(z'_t) \oplus f_{\mathrm{ext}}(e_t) \oplus \mathrm{Emb}(t) \bigr].
  \label{eq:reflex_input}
\end{equation}
Given an initial hidden state $h_0$ derived from the latent representation and the initial agent pose, the internal state then evolves via a causal temporal model (instantiated as a GRU in our implementation, though compatible with other causal sequence encoders),
\begin{equation}
  h_t = \mathcal{F}_{\theta}(u_t, h_{t-1}).
  \label{eq:reflex_gru}
\end{equation}
Finally, a generic decoder function $\mathcal{G}_{\phi}(\cdot)$ maps the internal state $h_t$ along with the current input $u_t$ to three execution primitives,
\begin{equation}
  \begin{bmatrix}x_{\mathrm{reflex}}\\ F_{\mathrm{ref}}\\ \Phi_K\end{bmatrix}
  = \mathcal{G}_{\phi}(h_t, u_t).
  \label{eq:reflex_decode}
\end{equation}
The decoder emits three coupled outputs: $x_{\mathrm{reflex}}$ matches the backbone pose format and, streamed each frame, can override the chunk after contact to limit geometric drift; $F_{\mathrm{ref}}$ is the desired contact wrench for sustained normal-pressure tracking; and $\Phi_K$ encodes end-effector stiffness and supports decoupling the direction and magnitude of applied effort.

\subsection{Contact-Triggered Phase Gating}
\label{subsec:phase_gating}

At deployment we alternate between two execution regimes consistent with our benchmarks.
\textbf{Phase~I:} before sustained contact, the low-level controller prioritizes tracking the frozen generative policy's pose targets.
\textbf{Phase~II:} once contact is detected from proprioceptive sensing, we switch emphasis to regulating $F_{\mathrm{ref}}$ and the impedance parameters implied by $\Phi_K$ through the VIC interface, while still streaming single-step $x_{\mathrm{reflex}}$ corrections for geometric drift.

%===============================================================================

\section{Experiments}
\label{sec:result}

We design experiments to answer six questions:
\textbf{(Q1)}~Can UniReflex improve contact-stage performance while preserving the backbone's non-contact visuomotor behavior?
\textbf{(Q2)}~How does UniReflex compare with end-to-end force-aware policies in position- and force-dominated execution?
\textbf{(Q3)}~Do learned force/stiffness outputs, high-frequency residuals, and contact-aware gating account for these gains?
\textbf{(Q4)}~Does the fast sub-network enable high-frequency recovery under environmental disturbances?
\textbf{(Q5)}~Does UniReflex improve force tracking and contact robustness during interaction?
\textbf{(Q6)}~Does its lightweight reflex design provide superior training throughput and memory efficiency over joint optimization?

%===============================================================================

\subsection{Setup}
\label{sec:exp_setup}

We evaluate five contact-rich benchmarks (Fig.~\ref{fig:tasks_overview}).
Our primary study couples UniReflex with three mainstream backbones: \textbf{DP}~\cite{chi2025diffusion}, $\boldsymbol{\pi}_{0}$~\cite{black2026pi0visionlanguageactionflowmodel}, and \textbf{DreamZero}~\cite{ye2026worldactionmodelszeroshot}.
In all cases, the generative stack remains frozen (i.e., already task-adapted) and only UniReflex is trained (Stage~II) in a plug-and-play manner without further slow-policy fine-tuning.
We additionally report force-aware monolithic baselines RDP~\cite{xue2025reactive}, ForceVLA~\cite{NEURIPS2025_8633b46e}, and TA-VLA~\cite{zhang2025elucidating} (Table~\ref{tab:eval_success_rates_stages}).
Demonstrations are collected via teleoperation at 200 trajectories per task.

%===============================================================================

\subsection{Results}
\label{sec:exp_results}

\begin{table*}[t]
  \centering
  \caption{\textbf{Success Rates (\%) on Contact-Rich Benchmarks.} Phase~I vs.\ Phase~II; UniReflex-augmented policies vs.\ vanilla backbones and force-aware baselines.}
  \label{tab:eval_success_rates_stages}
  \small
  % \tabcolsep=0 + in-cell \hspace avoids unpainted inter-column glue (white vertical
  % stripes) when using \rowcolor; outer @{} keeps the block flush to \textwidth.
  \begingroup
  \setlength{\tabcolsep}{0pt}
  \renewcommand{\arraystretch}{1.12}
  \begin{tabularx}{\textwidth}{@{}>{\raggedright\arraybackslash\hspace{2.8pt}}p{4.2cm}<{\hspace{2.8pt}} *{10}{>{\centering\arraybackslash\hspace{2.8pt}}X<{\hspace{2.8pt}}} @{}}
    \toprule
    \multirow{2}{*}{\textbf{Policy}}
      & \multicolumn{2}{c}{\textbf{Wipe (Curve)}}
      & \multicolumn{2}{c}{\textbf{Wipe (Incline)}}
      & \multicolumn{2}{c}{\textbf{Plug Charger}}
      & \multicolumn{2}{c}{\textbf{Peel Note}}
      & \multicolumn{2}{c}{\textbf{Flip Chip Box}} \\
    \cmidrule(lr){2-3} \cmidrule(lr){4-5} \cmidrule(lr){6-7} \cmidrule(lr){8-9} \cmidrule(lr){10-11}
      & \textbf{lift} & \textbf{wipe}
      & \textbf{lift} & \textbf{wipe}
      & \textbf{lift} & \textbf{plug}
      & \textbf{lift} & \textbf{peel}
      & \textbf{push} & \textbf{flip} \\
    \midrule
    RDP
      & 60.0\% & 50.0\% & 80.0\% & 70.0\% & 60.0\% & \textbf{50.0\%} & 60.0\% & 50.0\% & 70.0\% & 50.0\% \\
    ForceVLA
      & 30.0\% & 20.0\% & 50.0\% & 30.0\% & 30.0\% & 0.0\% & 50.0\% & 30.0\% & 40.0\% & 20.0\% \\
    TA-VLA
      & 50.0\% & 50.0\% & 70.0\% & 50.0\% & 40.0\% & 40.0\% & 70.0\% & 40.0\% & 50.0\% & 30.0\% \\
    \midrule
    DP
      & 70.0\% & 30.0\% & 80.0\% & 30.0\% & 70.0\% & \textbf{50.0\%} & \textbf{90.0\%} & 30.0\% & 80.0\% & 30.0\% \\
    $\pi_{0}$
      & \textbf{80.0\%} & 30.0\% & \textbf{90.0\%} & 30.0\% & \textbf{80.0\%} & \textbf{50.0\%} & \textbf{90.0\%} & 30.0\% & \textbf{90.0\%} & 30.0\% \\
    DreamZero
      & 60.0\% & 20.0\% & 60.0\% & 20.0\% & 50.0\% & 30.0\% & 70.0\% & 10.0\% & 70.0\% & 20.0\% \\
    \midrule
    \rowcolor{lightgreen}
    \textbf{UniReflex + DP}
      & \textbf{80.0\%} & 70.0\% & 80.0\% & 70.0\% & \textbf{80.0\%} & \textbf{50.0\%} & \textbf{90.0\%} & \textbf{90.0\%} & 80.0\% & 70.0\% \\
    \rowcolor{lightgreen}
    \textbf{UniReflex + $\boldsymbol{\pi}_{0}$}
      & \textbf{80.0\%} & \textbf{80.0\%} & \textbf{90.0\%} & \textbf{80.0\%} & 70.0\% & \textbf{50.0\%} & \textbf{90.0\%} & 80.0\% & 80.0\% & \textbf{80.0\%} \\
    \rowcolor{lightgreen}
    \textbf{UniReflex + DreamZero}
      & 60.0\% & 50.0\% & 70.0\% & 50.0\% & 60.0\% & 30.0\% & 60.0\% & 30.0\% & 60.0\% & 50.0\% \\
    \bottomrule
  \end{tabularx}
  \endgroup
\end{table*}

\textbf{Q1: Task Performance Across Execution Phases.}
For each backbone $X\in\{\mathrm{DP},\pi_{0},\mathrm{DreamZero}\}$, Table~\ref{tab:eval_success_rates_stages} compares UniReflex+$X$ with vanilla $X$.
We report task-level success over the two consecutive phases: Phase~II is attempted only after successful completion of Phase~I, and a Phase~I failure is therefore retained as a failure in the Phase~II success rate.
This end-to-end measure accounts for the approach and alignment behavior required before force-regulated execution can begin.
Across all tasks, adding UniReflex changes Phase~I success by no more than 10 percentage points relative to the corresponding backbone, indicating that visuomotor performance during the non-contact phase is largely preserved.
In Phase~II, UniReflex improves success by 20--60 percentage points on every sustained-contact benchmark across all three backbones.
Plug Charger is the only task without a clear improvement.
Unlike the other benchmarks, it is dominated by brief contact transients and tight insertion tolerances, which leave less scope for sustained force regulation.

\textbf{Q2: Force-Dominated Execution vs.\ Monolithic Baselines.}
Table~\ref{tab:eval_success_rates_stages} contrasts UniReflex+DP~\cite{chi2025diffusion}, UniReflex+$\pi_{0}$~\cite{black2026pi0visionlanguageactionflowmodel}, and UniReflex+DreamZero~\cite{ye2026worldactionmodelszeroshot} with monolithic RDP~\cite{xue2025reactive}, ForceVLA~\cite{NEURIPS2025_8633b46e}, and TA-VLA~\cite{zhang2025elucidating}.
On Phase~I, the frozen planner keeps the original backbone's position-control quality: UniReflex+DP matches or improves every alignment cell relative to vanilla DP, whereas the end-to-end force-aware models degrade on the same columns; specifically, ForceVLA and TA-VLA often trail by tens of points, and RDP is lower or tied on each lift/push slot. UniReflex+$\pi_{0}$ also outscores RDP, ForceVLA, and TA-VLA on every Phase~I cell.
On Phase~II, UniReflex+DP and UniReflex+$\pi_{0}$ meet or beat the best monolithic numbers.
Among the three force-aware policy baselines, RDP is the strongest; nevertheless, RDP still sits below our DP and $\pi_{0}$ entries on most contact-heavy cells.

\par\addvspace{5pt}
\noindent
\textbf{Q3: Component Contributions.}
\par\nobreak\addvspace{2pt}
\begin{wraptable}{r}{0.58\linewidth}
  \vspace{-8pt}
  \centering
  \caption{\textbf{Component ablations under frozen DP.}}
  \label{tab:component_ablation_archive}
  \resizebox{\linewidth}{!}{%
  \scriptsize
  \setlength{\tabcolsep}{3.5pt}
  \renewcommand{\arraystretch}{1.05}
  \begin{tabular}{@{}lccc@{}}
    \toprule
    \textbf{Variant} & \textbf{Peel Note} & \textbf{Wipe (Curve)} & \textbf{Wipe (Incline)}\\
    \midrule
    \multicolumn{4}{@{}l}{\textbf{Force/stiffness output (Phase~II success / F-err.)}}\\[-1pt]
    \rowcolor{lightgreen}
    \textbf{UniReflex (DP)} & \textbf{90.0\% / 5.8\%} & \textbf{70.0\% / 8.9\%} & \textbf{70.0\% / 6.8\%}\\
    No stiffness mod. & 50.0\% / 32.4\% & 30.0\% / 38.7\% & 40.0\% / 35.6\%\\
    Fixed isotropic stiffness & 50.0\% / 28.3\% & 40.0\% / 34.2\% & 30.0\% / 36.5\%\\
    Wrench-only & 30.0\% / 34.1\% & 20.0\% / 37.4\% & 20.0\% / 33.8\%\\
    Manual impedance & 60.0\% / 18.4\% & 50.0\% / 19.2\% & 40.0\% / 17.6\%\\
    \midrule
    \multicolumn{4}{@{}l}{\textbf{Residual/gating controls (Phase~I / Phase~II success)}}\\[-1pt]
    \rowcolor{lightgreen}
    \textbf{UniReflex (DP)} & \textbf{90.0\% / 90.0\%} & \textbf{80.0\% / 70.0\%} & \textbf{80.0\% / 70.0\%}\\
    Gated $\Delta x$ GRU & 90.0\% / 60.0\% & 80.0\% / 40.0\% & 80.0\% / 50.0\%\\
    Ungated GRU & 0.0\% / 0.0\% & 30.0\% / 0.0\% & 30.0\% / 10.0\%\\
    \bottomrule
  \end{tabular}%
  }
  \vspace{-8pt}
\end{wraptable}
\noindent
As shown in \mbox{Table~\ref{tab:component_ablation_archive}}, we conduct two ablation studies with a frozen DP backbone on Peel Note, Wipe (Curve), and Wipe (Incline) to explain the gains observed in Q1.

We test four alternative force/stiffness configurations: (i) disabling learned stiffness modulation while retaining the baseline stiffness; (ii) replacing the learned dynamic anisotropic stiffness with a manually selected isotropic stiffness that remains fixed during each task; (iii) retaining only reference-wrench prediction while removing the stiffness-direction output (wrench-only); and (iv) replacing state-dependent learned stiffness with a hand-designed controller using per-task fixed impedance schedules and manually calibrated gains. Alongside Phase~II success, we quantify contact-stage force tracking using
\begingroup
\setlength{\abovedisplayskip}{6pt}
\setlength{\belowdisplayskip}{6pt}
\setlength{\abovedisplayshortskip}{4pt}
\setlength{\belowdisplayshortskip}{6pt}
\begin{equation}
\label{eq:force_error}
\mathrm{F\mbox{-}err.} =
\frac{\sum_{t=1}^{T_{\mathrm{II}}}
  \left\lVert F^{\mathrm{meas}}_t-F^{\mathrm{tar}}_t \right\rVert_1}
{\sum_{t=1}^{T_{\mathrm{II}}}
  \left\lVert F^{\mathrm{tar}}_t \right\rVert_1},
\end{equation}
\endgroup
where $T_{\mathrm{II}}$ is the number of Phase~II samples and $F^{\mathrm{meas}}_t$ and $F^{\mathrm{tar}}_t$ are the measured and target force vectors; F-err.\ is reported as a percentage. Across the first three component-reduced configurations, Phase~II success drops from 70.0--90.0\% to 20.0--50.0\%, while F-err.\ rises from 5.8--8.9\% to 28.3--38.7\%. The fourth, hand-designed configuration performs better than these reduced variants, but remains below the complete learned formulation.

The second study separates the roles of stiffness modulation and contact gating. The gated $\Delta x$ GRU retains the gate but replaces learned stiffness modulation with $\Delta x$-based force regulation: it preserves 80.0--90.0\% Phase~I success, yet repeatedly adjusting $\Delta x$ to regulate contact force accumulates pose error under contact constraints, limiting Phase~II success to 40.0--60.0\% versus 70.0--90.0\% for full UniReflex. The ungated GRU instead removes the gate and relies on the fast network throughout both phases, reducing Phase~I/~II success to 0.0--30.0\%/0.0--10.0\%. Thus, $\Delta x$-based force regulation cannot replace learned stiffness modulation, while gating is essential for protecting the frozen planner before contact.

% Fig.~\ref{fig:reflex_force_tracking} (left) summarizes Peel Note and Wipe (Curve) under mild workspace disturbances and highlights how often the fast reflex updates relative to the frozen planner when those disturbances arrive.
\makeatletter
\setlength{\@fpsep}{6pt}% keep stacked top floats tight (must be outside local group)
\makeatother
\setlength{\textfloatsep}{8pt}
\begingroup
\setlength{\floatsep}{6pt plus 1pt minus 2pt}
\setlength{\intextsep}{8pt plus 2pt minus 2pt}
\begin{figure}[t]
  \centering
  \includegraphics[width=1.00\linewidth]{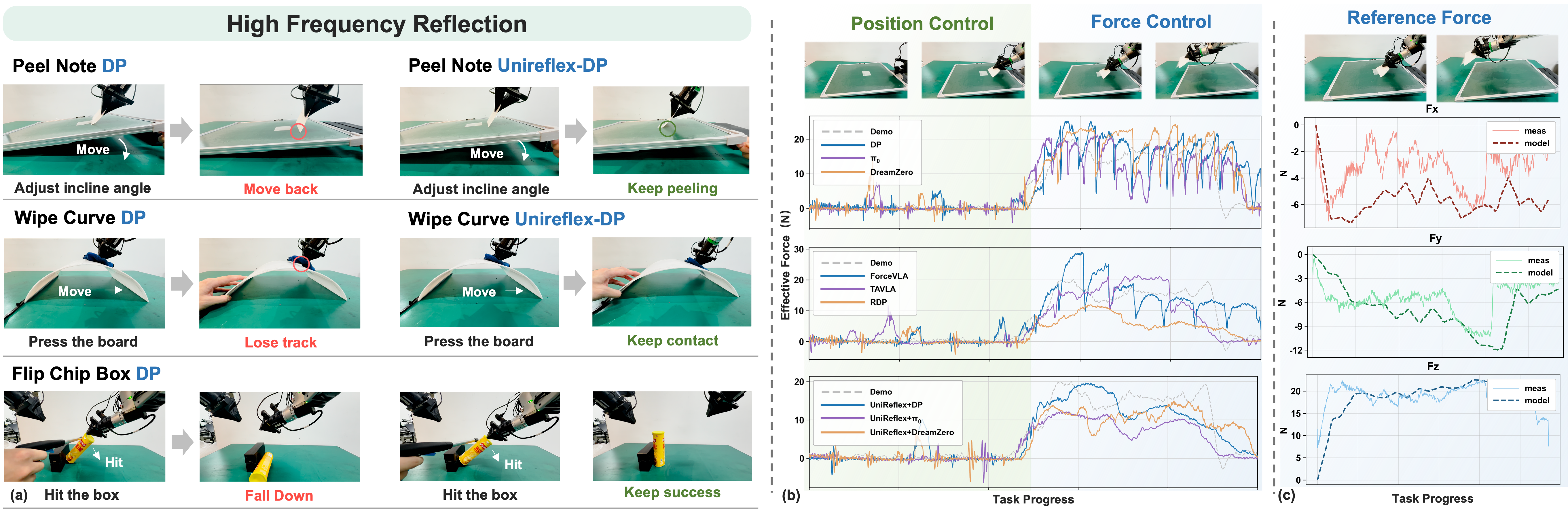}
  \caption{\textbf{High-frequency reflex and force tracking:}~\textbf{(a)} UniReflex maintains success on contact-rich tasks under disturbances. \textbf{(b)} Effective force during the peel task for baselines vs.\ UniReflex. \textbf{(c)} Measured contact force vs.\ predicted reference force during UniReflex execution.}
  \label{fig:reflex_force_tracking}
\end{figure}
\endgroup

\par\addvspace{5pt}
\noindent
\textbf{Q4: High-Frequency Recovery under Disturbances.}
\par\nobreak\addvspace{2pt}
\begin{wraptable}{r}{0.58\linewidth}
  \vspace{-8pt}
  \centering
  \caption{\textbf{Recovery under two perturbation settings.}}
  \label{tab:recovery_archive}
  \resizebox{\linewidth}{!}{%
  \setlength{\tabcolsep}{3.5pt}
  \renewcommand{\arraystretch}{1.05}
  \begin{tabular}{@{}lllccc@{}}
    \toprule
    \textbf{Perturbation} & \textbf{Policy} & \textbf{Metric} & \textbf{Peel} & \textbf{Wipe} & \textbf{Flip}\\
    \midrule
    \multirow{2}{*}{Dynamic} & DP & Rec. & 10.0\% & 20.0\% & 10.0\%\\
    & \cellcolor{lightgreen}\textbf{UniReflex (DP)} & \cellcolor{lightgreen}Rec./$T_{\mathrm{rec}}$ & \cellcolor{lightgreen}\textbf{90.0\%/0.87s} & \cellcolor{lightgreen}\textbf{100.0\%/1.13s} & \cellcolor{lightgreen}\textbf{80.0\%/1.28s}\\
    \midrule
    TCP bias (1--2\,cm) & \cellcolor{lightgreen}\textbf{UniReflex (DP)} & \cellcolor{lightgreen}Phase~II Rec. & \cellcolor{lightgreen}\textbf{80.0\%} & \cellcolor{lightgreen}\textbf{90.0\%} & \cellcolor{lightgreen}\textbf{70.0\%}\\
    \bottomrule
  \end{tabular}%
  }
  \vspace{-8pt}
\end{wraptable}
\noindent
To evaluate rapid reactions to online workspace changes, we test three tasks under dynamic disturbances, as shown in Fig.~\ref{fig:reflex_force_tracking}.
Specifically, we continuously adjust the support inclination during Peel Note, press the curved board during Wipe (Curve) to induce deformation, and perturb the chip box during Flip Chip Box to create unexpected contact changes.
As shown in Fig.~\ref{fig:reflex_force_tracking}(a), frozen DP loses contact under disturbances, whereas UniReflex+DP maintains reflex corrections and completes the tasks.

Table~\ref{tab:recovery_archive} reports 10 trials per task under two perturbations. Under dynamic disturbances, UniReflex recovers 80.0--100.0\% of executions and re-establishes the target contact force within 0.87--1.28s, versus 10.0--20.0\% recovery for frozen DP. With a 1--2\,cm pre-contact TCP bias, UniReflex recovers Phase~II execution in 70.0--90.0\% of trials, demonstrating compensation for bounded approach errors.

\par\vspace{4pt}
\noindent
\textbf{Q5: Interaction-Time Force Tracking.}
As shown in Fig.~\ref{fig:reflex_force_tracking}(b), UniReflex-augmented policies track demonstration force trajectories much more closely during the peeling task than frozen baselines.
This stability explains their higher Phase~II success in contact-rich tasks, especially when visual cues are sparse and force control is required to stabilize interaction.
Focusing on DP (Fig.~\ref{fig:reflex_force_tracking}(c)), UniReflex sustains peel contact by aligning measured 3D forces ($F_x$, $F_y$, $F_z$) with the predicted references, whereas frozen DP fails to keep forces within the contact-compatible band.
	
\textbf{Q6: Training Throughput and Memory Efficiency.}
As shown in Table~\ref{tab:speed_architecture_overview}, we compare estimated joint fine-tuning of the full stack with UniReflex Stage~II, which trains only the fast GRU, on a single NVIDIA A800 GPU.
Across the evaluated backbones, this reduces backward-pass latency by $25$--$66\times$ while training only 0.188--0.375\% as many parameters as the backbone.
The smaller trainable state also substantially reduces optimizer-state storage and activation traffic.

\begin{table}[H]
  \centering
  \caption{\textbf{Backbone scale vs.\ reflex training cost}. \textbf{Joint FT:} estimated full forward-backward cost; \textbf{Ours:} fast GRU backward step only.}
  \label{tab:speed_architecture_overview}
  \resizebox{0.94\columnwidth}{!}{%
  \footnotesize
  \setlength{\tabcolsep}{5pt}
  \renewcommand{\arraystretch}{1.05}
  \begin{tabular}{@{}lcccccc@{}}
    \toprule
    Backbone
      & Slow/Fast Params
      & Ratio
      & Steps $K$ / Size $H$
      & Joint FT (ms)
      & \textbf{Ours (ms)}
      & \textbf{Speedup}\\
    \midrule
    DP
      & 452.3M / 1.70M
      & 0.375\%
      & 20 (DDIM) / 256
      & $\approx$406
      & \textbf{7.70}
      & \textbf{$\approx$53$\times$} \\
    $\pi_{0}$
      & 3.50B / 6.60M
      & 0.188\%
      & 10 (Flow-Matching) / 1{,}024
      & $\approx$566
      & \textbf{8.57}
      & \textbf{$\approx$66$\times$} \\
    DreamZero
      & 22.92B / 62.2M
      & 0.271\%
      & 8 (Flow-Matching) / 5{,}120
      & $\approx$5{,}490
      & \textbf{223}
      & \textbf{$\approx$25$\times$} \\
    \bottomrule
  \end{tabular}%
  }
\end{table}

\section{Limitations}
\label{sec:limitations}

First, gains are smaller for long-horizon world-action backbones such as DreamZero, because their slowly refreshed action chunks limit timely responses to physical contact changes in the real world.
Second, UniReflex is better suited to sustained contact than brief, high-impulse rigid insertion, making tasks such as Plug Charger more challenging.
Third, achievable responsiveness remains bounded by force-sensing quality, low-level control, and hardware bandwidth.

% Acknowledgments appear only with [final] or [preprint] (hidden in anonymous review mode).
\acknowledgments{We thank X Square Robot and Xspark AI for their support of this research.}

%===============================================================================

\bibliography{unireflex}  % .bib

\end{document}